\documentclass[10pt,twocolumn,letterpaper]{article}

\usepackage{cvpr}              
\usepackage{pifont}

\usepackage{marvosym}
\definecolor{cvprblue}{rgb}{0.21,0.49,0.74}
\usepackage[pagebackref,breaklinks,colorlinks,allcolors=cvprblue]{hyperref}
\usepackage{adjustbox}

\usepackage{algorithm}
\usepackage{algorithmic}
\usepackage{multirow}
\usepackage{booktabs}
\usepackage{color}
\usepackage{tikz}
\usepackage{pifont}       
\usepackage{fontawesome}  
\usepackage{pifont}
\usepackage{graphicx}
\usepackage{overpic}
\usepackage{multirow}
\usepackage{booktabs}
\usepackage{color}
\usepackage{tikz}
\usepackage{adjustbox}  
\usepackage{amsmath}

\usepackage{wasysym}
\usepackage{amssymb}
\usepackage{multirow}
\usepackage{nicematrix}
\usepackage{booktabs}

\def\paperID{4193} 
\def\confName{CVPR}
\def\confYear{2026}

\title{UniShield: An Adaptive Multi-Agent Framework for Unified Forgery Image Detection and Localization}

\author{
Qing Huang\textsuperscript{1}, Zhipei Xu\textsuperscript{2}, Xuanyu Zhang\textsuperscript{2}, Xiangyu Yu\textsuperscript{1,3 \Letter}, Jian Zhang\textsuperscript{2,4}\\
\textsuperscript{1} School of Future Technology, South China University of Technology \\
\textsuperscript{2} School of Electronic and Computer Engineering, Peking University \\
\textsuperscript{3} School of Electronic and Information Engineering, South China University of Technology \\
\textsuperscript{4} Guangdong Provincial Key Laboratory of Ultra High Definition Immersive Media Technology, \\ Shenzhen Graduate School, Peking University \\ 
\vspace{-30pt}
}

\begin{document}

\twocolumn[{
\renewcommand\twocolumn[1][]{#1}
\maketitle
\vspace{-5mm}
\begin{center}
    \captionsetup{type=figure}
    \includegraphics[width=0.95\linewidth]{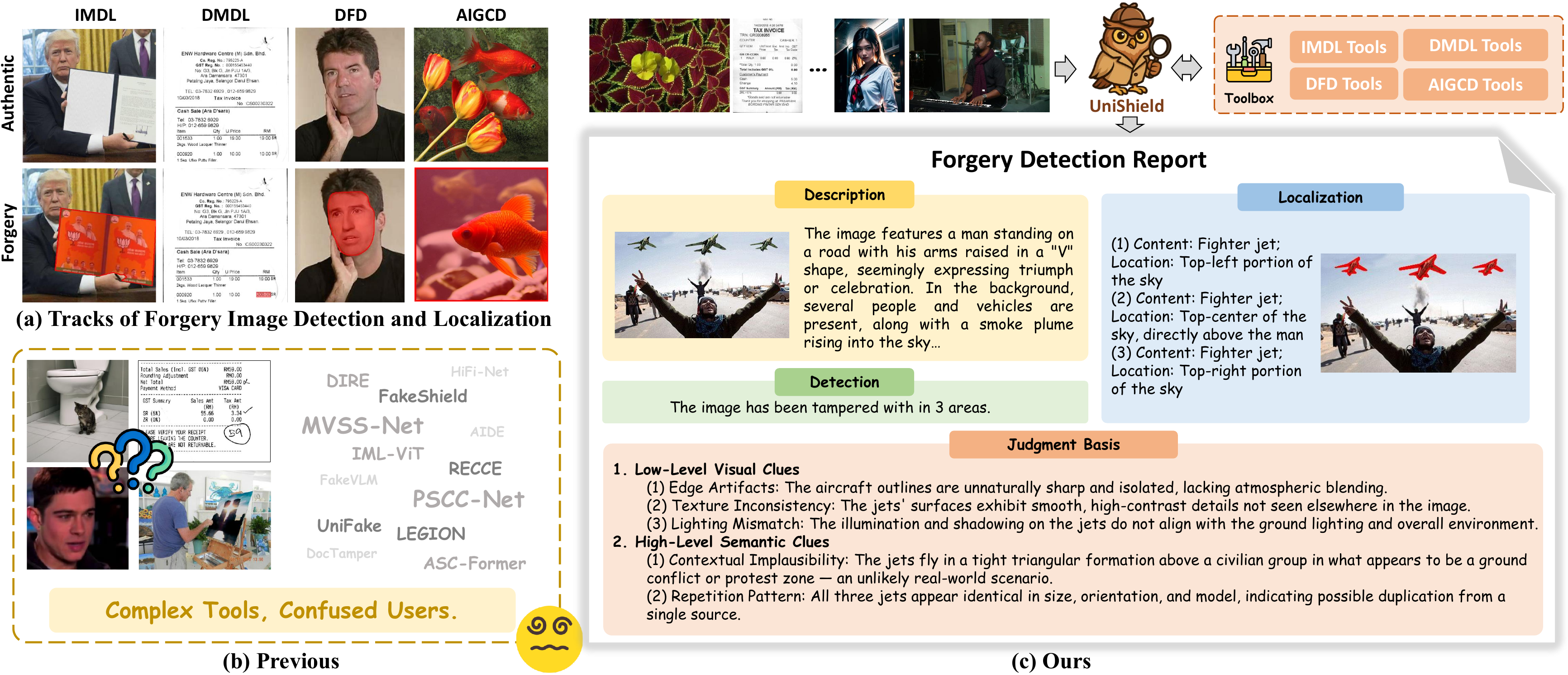}
    \vspace{-8pt}
    \captionof{figure}{ \textbf{Unified Framework for Forgery Image Detection and Localization.} (a) Various types of forgery images (b) Previous methods required users to manually determine the detection tools based on image content, often leading to confusion. (c) Our system, UniShield, automatically coordinates various detection tools for efficient, unified forgery detection, and outputs interpretable reports including description, detection, localization, and judgment basis. }
    \label{teasor} 
\end{center}
\vspace{-5pt}
}]

\begin{abstract}

With the rapid advancements in image generation, synthetic images have become increasingly realistic, posing significant societal risks, such as misinformation and fraud. Forgery Image Detection and Localization (FIDL) thus emerges as essential for maintaining information integrity and societal security. Despite impressive performances by existing domain-specific detection methods, their practical applicability remains limited, primarily due to their narrow specialization, poor cross-domain generalization, and the absence of an integrated adaptive framework. To address these issues, we propose UniShield, the novel multi-agent-based unified system capable of detecting and localizing image forgeries across diverse domains, including image manipulation, document manipulation, DeepFake, and AI-generated images. UniShield innovatively integrates a perception agent with a detection agent. The perception agent intelligently analyzes image features to dynamically select suitable detection models, while the detection agent consolidates various expert detectors into a unified framework and generates interpretable reports. Extensive experiments show that UniShield achieves state-of-the-art results, surpassing both existing unified approaches and domain-specific detectors, highlighting its superior practicality, adaptiveness, and scalability.

\renewcommand\thefootnote{\relax} \footnote{\Letter: Corresponding author.}
\end{abstract}    

\section{Introduction}

In recent years, with the rapid advancement of deep learning~\cite{chen2024confusion,li2026comprehensive,yu2024identifying,zhou2024adversarial,zhang2023art} and generative technologies~\cite{wu2025qwen,zhang2024synergistic,zhang2024learnable,hou2026toward}, the quality and complexity of synthetic images have significantly improved, making them indistinguishable from real ones to the human. 
Although these tools can bring great convenience to creators and promote the development of the cultural industry, they are sometimes misused and may bring social problems such as the spread of fake news and online fraud.
As a result, in the era of AIGC, Forgery Image Detection and Localization (FIDL) has become a crucial technical task for ensuring information security and societal stability.

Current FIDL research~\cite{wang2023dire,huang2024ffaa,yu2024semgir,tan2025c2p,zhu2024hiding,yan2024effort,luo2024forgery} covers a wide range of manipulation types, and achieves impressive performance within their respective targeted domains.
While existing detection tools each have their own strengths across different manipulation types, their practical utility is hindered in real-world scenarios. Users often lack sufficient prior knowledge to select the most appropriate detection tool for a given image, as this requires not only a comprehensive understanding of the strengths and limitations of all available detectors, but also the ability to accurately anticipate which type of forgery the image is most likely subjected to. Therefore, a unified FIDL method is highly necessary to bridge this gap and enable more practical, automated FIDL in the wild.

However, achieving a unified FIDL framework still faces two major challenges. \textbf{Firstly}, models trained on mixed forgery datasets often experience performance degradation caused by domain conflicts. Conversely, models trained exclusively on a single forgery type exhibit strong in-domain performance but struggle to generalize across diverse domains. \textbf{Secondly}, existing forgery detection models tend to focus on isolated aspects of image features. Some methods emphasize frequency-domain artifacts~\cite{jeong2022frepgan}, while others rely primarily on spatial-domain cues~\cite{cao2022end}. Certain approaches are tailored to capture high-level semantic inconsistencies~\cite{kang2025legion,wen2025spot,sun2025towards}, whereas others specialize in detecting low-level traces such as noise or compression artifacts~\cite{ma2023iml,wang2025opensdi,yan2024sanity,liu2022pscc}. Despite this diversity, there is currently no effective mechanism to integrate these complementary detection strategies into a unified, more robust framework. As a result, existing methods face not only limitations when confronted with multi-domain forgeries but also a lack of a cohesive and adaptive scheduling strategy to fully leverage their potential.


Agent is an intelligent entity equipped with the capabilities of perception, reasoning, planning, and execution, enabling it to leverage various external tools to autonomously accomplish complex tasks. In recent years, agent-based frameworks have been widely adopted in various fields~\cite{liu2024medcot,xia2025mmedagent,wang2024videoagent,kumar2024mmctagent,yang2024agentoccam}, including image generation, multimodal understanding, and autonomous driving, demonstrating strong adaptability and generalization capabilities. However, the agent-based framework in the field of forgery detection remains unexplored.
Compared to traditional static models, the agent architecture can integrate expert tools from multiple FIDL tracks to fulfill unified detection requirements, which better aligns with real-world demands for generalization. In addition, agent systems offer greater flexibility and scalability, enabling the rapid integration of new forgery detection modules and effectively responding to continuously evolving forgery techniques.


To address these challenges, we present UniShield, the first multi-agent based system that offers a unified, scalable, and cross-domain adaptive solution for forgery image detection and localization.
As illustrated in Figure~\ref{teasor}, We observe that the FIDL task can be decomposed into four relatively orthogonal tracks: Image Manipulation Detection and Localization (IMDL), Document Manipulation Detection and Localization (DMDL), DeepFake Image Detection (DFD), and AIGC Image Detection (AIGCD). To handle this categorization, UniShield is equipped with two collaborative core components: the perception agent and the detection agent.
The perception agent analyzes the input image based on its semantic structure and low-level visual features, infers the likely type of forgery, and dynamically selects the most suitable detection model from a toolbox. Meanwhile, the detection agent aggregates expert models across all forgery domains into a comprehensive detector toolbox and outputs a structured, interpretable report upon completion, enhancing both the practicality and user-friendliness of forgery detection. Our contributions are summarized as follows:

\vspace{1pt}
\noindent \ding{113}~(1) We propose a novel multi-agent based forgery image detection and localization framework, UniShield. It can process suspected images on all domains, including IMDL, DMDL, DFD and AIGCD, and generate a detection report with strong interpretability, significantly improving the practicality and versatility of existing FIDL approaches.

\vspace{1pt}
\noindent \ding{113}~(2) We design a perception agent with a task router and a tool scheduler. The task router selects different detection tracks based on the image distribution, while the tool scheduler adaptively chooses the most suitable expert detector.

\vspace{1pt}
\noindent \ding{113}~(3) We design a detection agent that can efficiently integrate all non-LLM-based and LLM-based detector across different domains, while producing interpretable and insightful detection analyses.

\vspace{1pt}
\noindent \ding{113}~(4) Experiments show that our UniShield can achieve state-of-the-art performance across multiple benchmarks. UniShield outperforms existing all-domain detectors as well as all domain-specific expert detectors.

\begin{figure*}[t]
	\centering
	\includegraphics[width=1\linewidth]{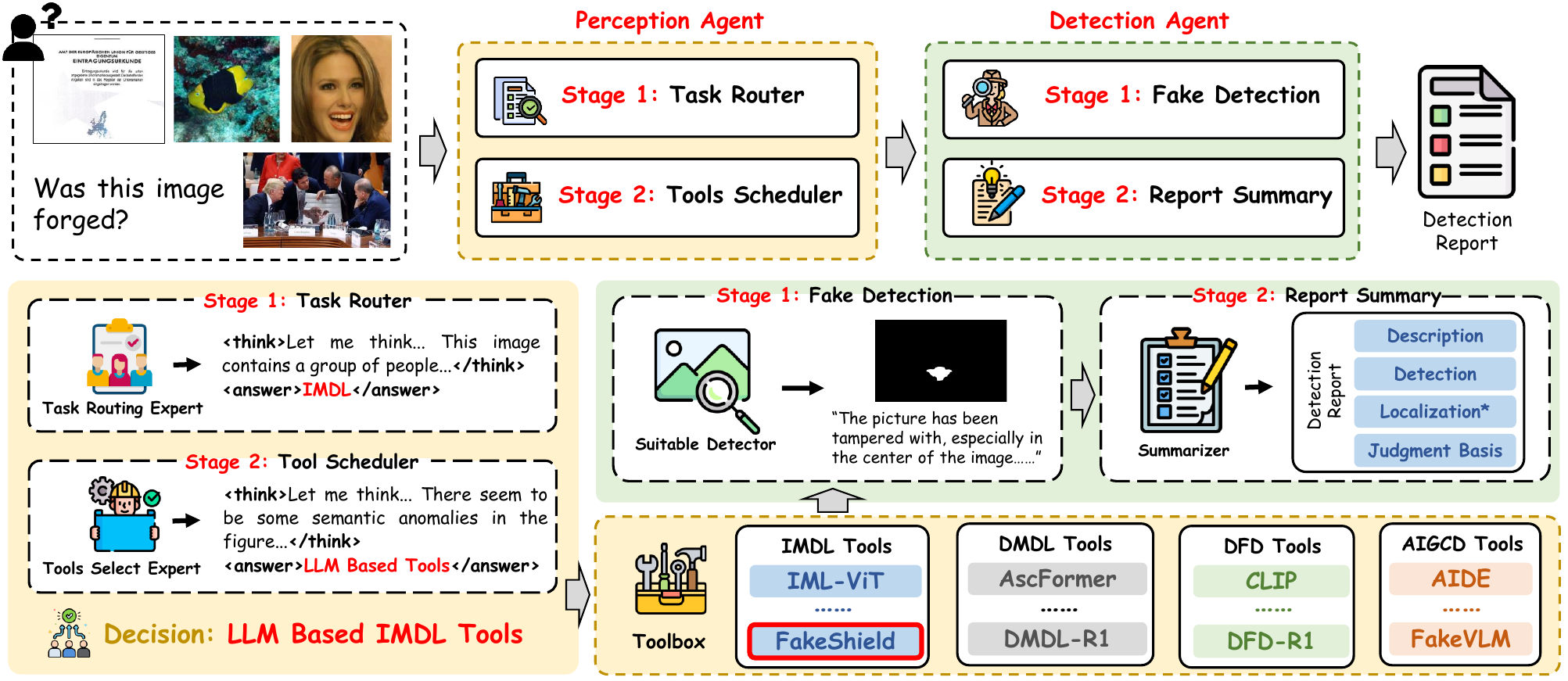}
	\vspace{-15pt}
	\caption{\textbf{The pipeline of UniShield.} Our method consists of two main components: Perception Agent and Detection Agent. The Perception Agent includes a task router that determines the forgery domain and a tool scheduler that selects the appropriate detector type based on image content. The Detection Agent then performs fake detection using the selected expert tool and generates a structured report, including detection result and the reasoning behind the judgment. }
	\label{pipeline}
    \vspace{-15pt}
\end{figure*}

\section{Related Works}
\subsection{Forgery Image Detection and Localization}



Current FIDL methods~\cite{zhong2023patchcraft,zhong2023rich,zhang2022patch,fang2024uniforensics,salvi2023robust,li2025texture,yang2025all,yan2024generalizing,zhou2024freqblender,qu2024textsleuth,cao2022end,zhang2024editguard,zhang2025omniguard} can be broadly categorized into four domains, with each domain encompassing two distinct approaches: large language model (LLM) based detection methods and non-LLM-based methods.

For IMDL, MVSS-Net~\cite{dong2022mvss} introduced multi-view and multi-scale supervision to jointly capture tampering boundaries and noise patterns. PSCC-Net~\cite{liu2022pscc} employed a progressive two-path framework with spatio-channel correlation to detect and localize manipulations across multiple scales with coarse-to-fine accuracy. FakeShield~\cite{xu2024fakeshield} introduced a vision-language model to detect and interpret image manipulations through cross-modal grounding. For DeepFake detection, VLFFD~\cite{sun2025towards} introduced a visual-linguistic paradigm that generates fine-grained sentence-level prompts to enhance deepfake detection interpretability and generalization via multimodal co-training. FakeFormer~\cite{nguyen2024fakeformer} enhanced vision transformers with artifact-guided attention to effectively capture subtle deepfake inconsistencies for improved generalization and efficiency.
For AI-generated image detection, AIDE~\cite{yan2024sanity} detected AI-generated images by combining CLIP-based semantic understanding with low-level artifact analysis through frequency-based and patch-based hybrid features. OpenSDI~\cite{wang2025opensdi} introduced a benchmark and the MaskCLIP framework to detect and localize diffusion-generated images in open-world scenarios via synergizing multiple pretrained models. For DMDL, DocTamper~\cite{qu2023towards} introduced a fine-grained framework and large-scale dataset for tampered text detection in document images, leveraging frequency-aware features and multi-view decoding; AscFormer~\cite{luo2024toward} introduced a dual-stream framework with consistency-aware aggregation and contrastive learning to effectively detect real-world tampered text using the RTM dataset. However, while these models can achieve SoTA results in their respective domains, they cannot be generalized to other domains, and no method integrates them. Recently, ForensicHub~\cite{du2025forensichub} further moves toward this goal by providing a unified benchmark and codebase for all-domain fake image detection and localization.

\subsection{LLM Agents}
The rapid advancement of large language models (LLMs) has driven the emergence of powerful multimodal systems~\cite{chen2023sharegpt4v,dai2023instructblip,zeng2025FSDrive,zhouboosting,xiaoreversible,zhangpi,liu2024fedbcgd,li2025frequency,li2025sepprune,zhang2025vq,zhao2025reasoning,xu2025avatarshield,chen2023bianque}. Equipped with broad world knowledge, strong reasoning capabilities, and the ability to interpret complex multimodal inputs, these models provide a solid foundation for building intelligent agents capable of structured decision-making and autonomous task execution. As a result, LLM-based agents have been successfully adapted to various domains.

In the medical field, MedCoT~\cite{liu2024medcot} acted as a hierarchical expert agent for medical reasoning; MMedAgent-RL~\cite{xia2025mmedagent} served as a collaborative multi-agent system for adaptive clinical diagnosis.
In visual comprehension, VideoAgent~\cite{wang2024videoagent} served as a central agent to extract and reason over key visual information in long-form videos; MMCTAgent~\cite{kumar2024mmctagent} acted as a critical-thinking agent that decomposes and verifies complex multimodal inputs for enhanced visual understanding.
In the area of image restoration and enhancement, MAIR~\cite{jiang2025multi} adopted a collaborative multi-agent design to address complex real-world degradations; 4KAgent~\cite{zuo20254kagent} integrated specialized perception and restoration agents to upscale low-quality images into high-fidelity 4K outputs.
In the field of image generation, Idea2Img~\cite{yang2024idea2img} acted as a self-refining agent that iteratively improves text-to-image generation through multimodal feedback; GenArtist~\cite{wang2024genartist} served as a unified MLLM agent that orchestrates tool selection, editing, and verification for high-quality image synthesis.
Although the agent framework has achieved success in many fields, it remains unexplored in image forgery detection, where no unified solution exists for diverse forgery cases. 

\section{Methodology}
\subsection{Overview}
We present UniShield, a general and unified forgery image detection and localization system that is capable of handling forged images across all domains.
As illustrated in Figure~\ref{pipeline}, UniShield is composed of two main modules: perception agent and detection agent, together forming an end-to-end framework for automatic FIDL.

When an image with unknown forgery characteristics is fed into the system, the perception agent first conducts a preliminary analysis. The task router analyzes the overall visual distribution of the image and outputs the image forgery attributes. Then, the tool scheduler determines whether to apply an LLM-based or non-LLM-based detection tool, depending on the image content. LLM-based methods are better suited for detecting semantic-level inconsistencies, while non-LLM-based models are more sensitive to low-level cues.
Once the appropriate tool is selected, the detection agent performs the analysis. Guided by the perception agent's decision, the system invokes the most suitable expert model from the toolbox. Upon completing the detection, the results are passed to the summarizer, which generates a structured report detailing the image content, tampered regions, and the reasoning behind the detection. This process enhances system interpretability and deepens users' understanding of the results. 
Noted that, for each image, we select only one detector from a single track for detection. Although some agent-based methods emphasize the importance of backtracking or multi-tool collaboration, we intentionally avoid such strategies to prevent potential conflicts between different tools.

\subsection{Perception Agent}

\begin{figure}[t]
	\centering
	\includegraphics[width=1.0\linewidth]{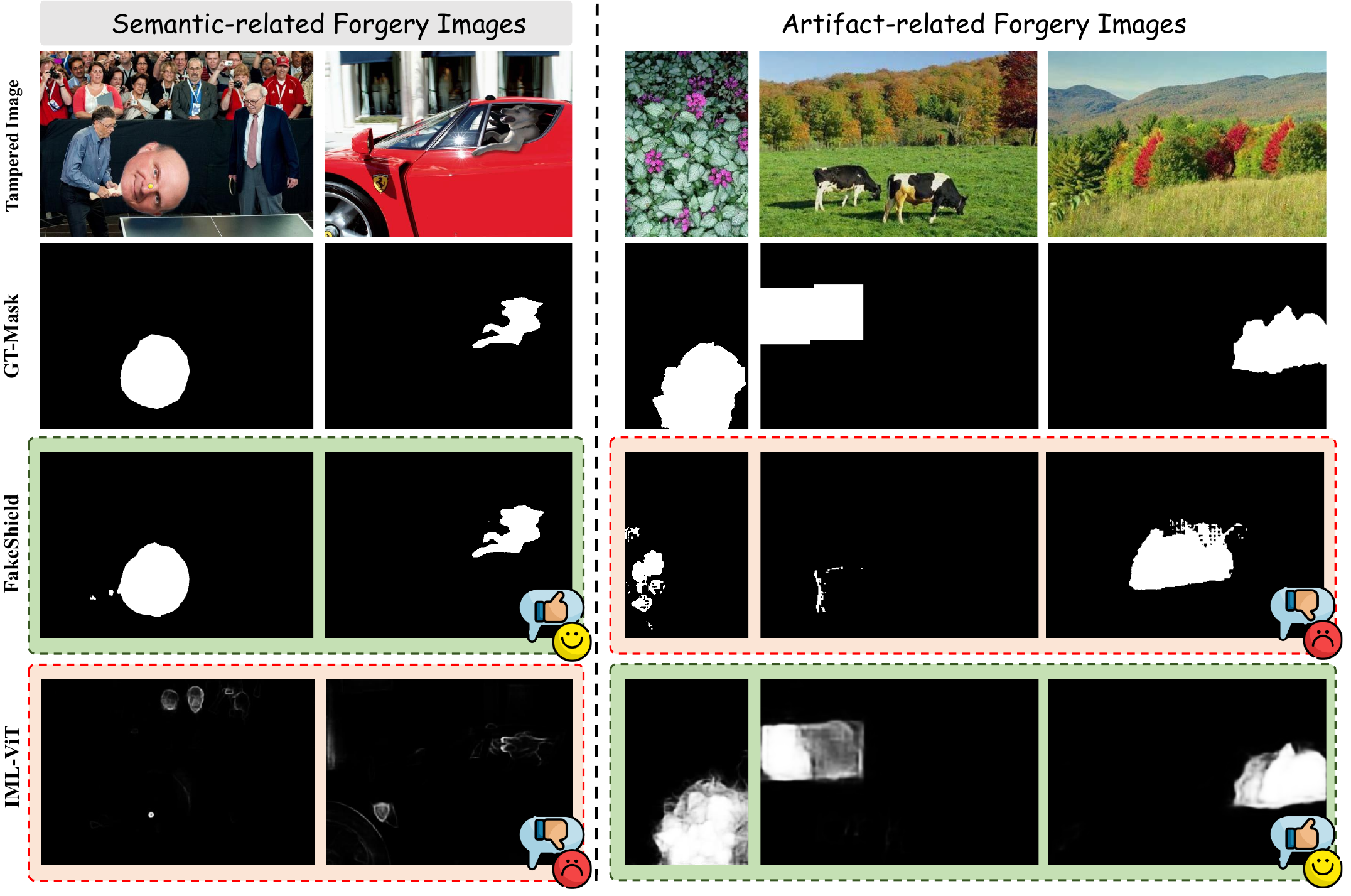}
	\vspace{-20pt}
	\caption{\textbf{Pilot study comparing LLM-based and non-LLM-based detectors on different types of forgeries.} The LLM-based model FakeShield performs better on semantic forgeries, while the non-LLM-based method IML-ViT excels at detecting low-level artifact-based manipulations.}
	\label{xiandao}
    \vspace{-15pt}
\end{figure}

\subsubsection{Task Router} 





Through our observation and research, current image forgery methods can be broadly classified into two categories: global forgery and local forgery. In global forgery, AI-generated images are the main focus. These forged images are generally created using techniques such as Diffusion and GAN~\cite{rombach2021highresolution} generating entirely fake images from latent features. Local forgery can be divided into three subcategories. The first is a series of DeepFake techniques~\cite{faceapp2017} focused on faces. These techniques rely on GANs and enable facial editing tasks like face swapping and face reenactment on real face images. The second subcategory involves local manipulation of general images using ``cheapfake'' techniques such as copy-move, splicing, removal, and inpainting, allowing for quick local edits. The third subcategory pertains to document image forgery, which is a special form of cheapfake. Due to the high information density of document images, we consider them separately. In summary, image forgery methods can be divided into four distinct categories, leading to four corresponding detection tracks: AIGCD, DFD, IMDL, and DMDL. The first two tasks focus on detection results, while the latter two require both detection and localization.

We aim to design an agent that can assign an image to its corresponding track, facilitating the selection of an appropriate forgery detector in subsequent steps. While images across different tracks may vary in content and data domains, the more critical distinction lies in the manipulation and forgery techniques applied. Therefore, this is not merely a standard image classification problem; it requires the router to possess a certain level of source tracing and forgery detection capability. This poses a significant challenge for conventional models with limited parameter sizes, prompting us to consider using a multimodal large language model (MLLM) to address the task. Although some pretrained MLLMs are equipped with broad world knowledge, this task clearly exceeds their inherent capabilities, necessitating fine-tuning of the MLLM. Notably, due to the scarcity of supervised fine-tuning (SFT) data and the need for interpretability of the entire process, we propose optimizing the MLLM using outcome-based reinforcement learning in the R1 style. Group Relative Policy Optimization (GRPO)~\cite{guo2025deepseek} is an advanced reinforcement learning algorithm derived from Proximal Policy Optimization (PPO). Unlike PPO, which relies on value estimation via a critic, GRPO evaluates multiple candidate responses collectively by comparing their relative rewards, thereby avoiding explicit value modeling. This design improves training stability and efficiency, making it particularly suitable for tasks with limited supervision or where outcome quality is best assessed relatively.
The optimization objective of GRPO is defined as:
\vspace{-20pt}

\begin{align}
\label{GRPO}
&\max_{\pi_\theta} \;  \mathbb{E}_{o \sim \pi_\theta(q)} [ R_{\text{GRPO}}(q, o) ] \\
&=[ R_{\text{total}}(q, o) - \beta\cdot\mathrm{KL}[ \pi_\theta(o|q)~\| ~\pi_{\text{ref}}(o|q)]],
\end{align}
where $\pi_{\text{ref}}$ is the reference model prior to optimization, $R_{\text{GRPO}}$ is the relative reward function used to compare candidates, $\beta$ is the hyperparameter controlling the KL divergence regularization, and $R_{\text{total}}(q, o)$ is the relative reward function that evaluates candidate outputs. The reward is defined as:
\vspace{-15pt}

\begin{equation}
R_{\text{total}}(q, o) = R_{\text{task}}(q, o)+R_{\text{format}}(q,o), \end{equation}
\begin{equation}
R^{(i)}_{\text{task}}(q,o) = 1, \quad \text{if } o^{(i)} = \text{task}_{\text{gt}}~\text{ else } 0,
\end{equation}
where $R_{\text{task}}$ is the task classification reward, and $R_{\text{format}}$ is the formatting reward. We select mainstream datasets from AIGCD, DFD, IMDL, and DMDL respectively, with each dataset domain containing both authentic and forged images. The overall workflow of the task router is illustrated in Figure~\ref{pipeline}. Specifically, we input an image into the Task Router and extract the content within the $<$answer$>$...$<$/answer$>$ tags. The content inside the answer will include one of the four labels: AIGCD, DFD, IMDL, or DMDL, indicating which type of forgery tool the agent believes most likely forgery the image.

\subsubsection{Tool Scheduler} Within each track, researchers have proposed both LLM-based and non-LLM-based methods. We observe that due to the world knowledge encoded during pretraining, LLM-based models are inherently more sensitive to semantic inconsistencies, while non-LLM-based methods are more proficient at identifying local visual artifacts.
We take the IMDL domain as an example and demonstrate this through a simple pilot experiment. We compare two well-known methods: FakeShield (LLM-based) and IML-ViT (non-LLM-based). As shown in Figure~\ref{xiandao}, the left two images are semantically manipulated, where the forged regions can be identified through logical reasoning. The right three images contain subtle visual artifacts that are difficult to spot with the naked eye. The results indicate that FakeShield performs better on semantically forged images, while IML-ViT excels at detecting artifact-based tampering. 



Therefore, we plan to use a module to initially identify whether an image contains logical or semantic inconsistencies, in order to guide the selection of a more suitable forgery detection tool. Considering the strong image content perception and logical reasoning capabilities of MLLMs, we employ Qwen2.5-VL~\cite{Qwen2.5-VL} as our tool scheduler.
We carefully design a set of prompts for Qwen2.5-VL to guide it in analyzing both the semantic structure and low-level visual features of an input image, enabling it to make a binary decision between “LLM-based” and “non-LLM-based” detection methods without any additional training. The prompt instructs the model to consider two key perspectives: (1)  If the image exhibits high-level semantic or logical inconsistencies, such as implausible object relationships, violations of commonsense, or contradictions in contextual elements, the model is directed to choose an LLM-based detection tool; (2) if the image displays low-level visual artifacts like texture discontinuities, edge anomalies, or compression traces, a non-LLM-based method is selected. The full prompt is provided in the supplementary materials.

By combining the results from the task router and the tool scheduler, the system identifies both the forgery domain and the appropriate model type. This joint output is then passed to the detection agent, which selects a suitable expert detector from the tool box accordingly. 


\subsection{Detection Agent}

\subsubsection{Fake Detection} To support comprehensive and accurate FIDL, UniShield incorporates a detector toolbox. Table~\ref{toolbox} presents the detection models integrated by UniShield. For the IMDL and DMDL tasks, in addition to forgery confidence scores, the models are required to output tampering region masks. Furthermore, LLM-based models provide natural language explanations that describe the nature of the forgery, enhancing the interpretability and user-friendliness of the detection results.


\begin{figure*}[t]
	\centering
	\includegraphics[width=1.0\linewidth]{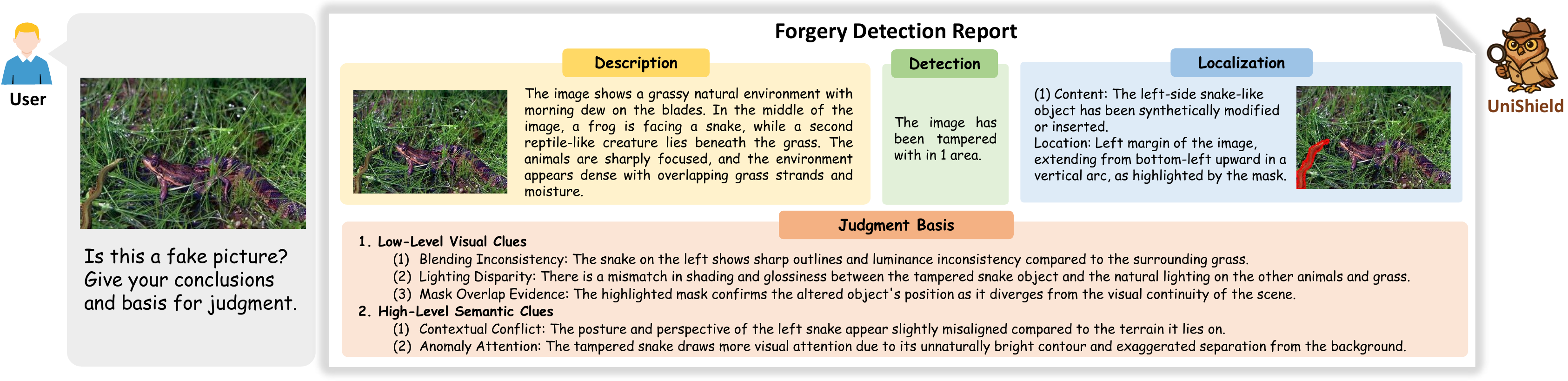}
	\vspace{-18pt}
	\caption{Illustration of the forgery report of our UniShield.}
	\label{zhuguan}
    \vspace{-15pt}
\end{figure*}

\begin{table}[t!]
\centering
\caption{The list of tools in the detector box.}
\vspace{-8pt}
\label{toolbox}
\renewcommand{\arraystretch}{1.2}
\setlength{\tabcolsep}{2pt} 
\resizebox{1.0\linewidth}{!}{
\begin{NiceTabular}{c|cc|cc}
\CodeBefore
\tikz \fill [gray!10] (1-|1) rectangle (2-|6);   
\tikz \fill [gray!10] (2-|1) rectangle (3-|6); 
\Body
\toprule[1pt]
\multirow{2}{*}{Sub-Domain} & \multicolumn{2}{c|}{non-LLM-based} & \multicolumn{2}{c}{LLM-based} \\
                            & model name & output & model name & output \\
\midrule
\multirow{2}{*}{IMDL}   & IML-ViT   & confidence,   & FakeShield & detection result, \\
                        & ~\cite{ma2023iml} & mask & ~\cite{xu2024fakeshield} & mask, explanation \\
\midrule
\multirow{2}{*}{DMDL}   & AscFormer & confidence,   & DMDL-R1    & detection result,  \\
                        & ~\cite{luo2024toward} & mask & (Ours) & mask, explanation \\ 
\midrule
\multirow{2}{*}{DFD}    & CLIP   & \multirow{2}{*}{confidence}  & DFD-R1 & detection result,     \\
                        & ~\cite{radford2021learning} &  & (Ours)  & explanation \\
\midrule
\multirow{2}{*}{AIGCD}  & AIDE      & \multirow{2}{*}{confidence}  & FakeVLM    & detection result,        \\
                        & ~\cite{yan2024sanity} &  & ~\cite{wen2025spot} & explanation \\
\bottomrule[1pt]
\end{NiceTabular}
}
\vspace{-10pt}
\end{table}

All the above detection tools are used by the tool scheduler to dynamically schedule calls based on the decision results of the perception agent. This design enables UniShield to maintain high flexibility and scalability, laying a solid foundation for future expert model integration. Notably, in the domains of DFD and DMDL, there are currently no open-source LLM-based approaches that meet our requirements. Due to the limited availability of SFT data and the need for interpretability, we consider using the GRPO to fine-tune Qwen-2.5VL for the tasks of DeepFake detection and document manipulation detection. Similar to the training of the task router, we optimize Qwen-2.5VL using Equation~\ref{GRPO}, where the reward function is defined as follows:
\vspace{-20pt}

\begin{equation}
R_{\text{total}}(q, o) = R_{\text{acc}}(q, o)+R_{\text{format}}(q,o).
\end{equation}

Here, $R_{\text{acc}}(q, o)$ is set to 1 when the model makes a correct detection, and 0 otherwise.  Additionally, since the DMDL task requires localization of forged regions, following the approach of FakeShield, we feed the textual output of the MLLM into GLaMM~\cite{rasheed2024glamm} and fine-tune it using the corresponding mask as ground truth.



\subsubsection{Report Summary} After the fake detection stage is complete, all detection results and the intermediate outputs from each step of the framework are fed into the Summarizer module. This component generates a structured and interpretable forgery detection report to help users clearly understand the detection results. We implement it using GPT-4o~\cite{openai2023gpt}, an SOTA MLLM known for producing fluent, context-aware, and highly interpretable summaries. To ensure consistency, we design a standardized vision-language prompt that guides the report generation process in a consistent format. The report includes three main parts: a brief description of the image content, a detection conclusion, and the reasoning behind the decision. The description summarizes the scene or subject in the image; the detection conclusion states whether the image is real or forged, and if the task needs to be located, you also need to output a mask-related description and mark the visible tampering area. The judgment basis is derived from the model’s output features, presenting low-level visual cues (e.g., unnatural textures, edge artifacts, inconsistent noise) or high-level semantic inconsistencies (e.g., unnatural facial expressions, scene conflicts). This module significantly enhances the interpretability and readability of the detection results, making the system more user-friendly and better suited for real-world deployment in security and forensic applications.

\section{Experiment}
\subsection{Experiment Setup}
\textbf{Implementation Details.}
Our UniShield framework integrates eight expert detectors across four major forgery domains, with the specific tools used detailed in Table~\ref{toolbox}. 
As for the agent modules, both the task router and tool scheduler are initialized using Qwen2.5-VL, and the report summarization is powered by GPT-4o. We apply GRPO to optimize the task router, DMDL-R1, and DFD-R1, and perform full-parameter fine-tuning using the R1-V framework~\cite{chen2025r1v}.
The training is conducted using 4 NVIDIA A800 80GB GPUs, with a learning rate set to $1$$\times$$10^{-6}$, and runs for one epoch. For GRPO-based training, we set the $\beta$ coefficient to 0.04 to regulate the divergence between the policy model and the reference model during optimization. Further details can be found in the supplementary material.

\textbf{Test Datasets.}
To comprehensively evaluate the performance of UniShield across diverse forgery types, we conduct experiments on multiple benchmark datasets covering all four domains. 
For IMDL, we use CASIA1+~\cite{dong2013casia} and IMD2020~\cite{novozamsky2020imd2020}, which provide pixel-level annotations for tampered natural images. For DMDL, we evaluate on RTM~\cite{luo2024toward}, a benchmark targeting realistic document forgeries. For AIGC detection, we adopt AIGCDetectionBenchmark~\cite{zhong2023patchcraft}, covering a wide range of generative models and synthetic content. For DeepFake detection, we use DF40~\cite{yan2024df40}, a large-scale benchmark with 40 diverse face forgery techniques.

\begin{table}[t]
\centering
\caption{Cross-domain detection performance. The best score in each column is highlighted in \textbf{bold} and the second-best score is \underline{underlined}.}
\vspace{-8pt}
\label{table:cross}
\renewcommand{\arraystretch}{1.2}
\resizebox{1.0\linewidth}{!}{  
\begin{NiceTabular}{c|cc|cc|cc|cc}
\CodeBefore
\tikz \fill [gray!10] (1-|1) rectangle (2-|10);   
\tikz \fill [gray!10] (2-|1) rectangle (3-|10); 
\Body
\toprule[1pt]
\multirow{2}{*}{Method} & \multicolumn{2}{c|}{IMDL} & \multicolumn{2}{c|}{AIGCD} & \multicolumn{2}{c|}{DFD} & \multicolumn{2}{c}{DMDL} \\
& ACC & F1 & ACC & F1 & ACC & F1 & ACC & F1 \\
\midrule
ResNet      & 0.840  & 0.827  & 0.765  & \underline{0.774}  & 0.141  & 0.214  & \underline{0.629}  & \underline{0.650} \\
CLIP        & 0.654 & 0.533 & 0.659  & 0.647  & 0.278  & 0.415  & 0.618  & 0.625 \\
HiFi-Net    & 0.464  & 0.449  & 0.678  & 0.664  & 0.428  & 0.451  & 0.118  & 0.104 \\
FakeShield  & \underline{0.945}  & \underline{0.948}  & \underline{0.781}  & 0.754  & \underline{0.675}  & \underline{0.710}  & 0.147  & 0.162 \\
Ours        & \textbf{0.971}  & \textbf{0.966}  & \textbf{0.942}  & \textbf{0.931}  & \textbf{0.913}  & \textbf{0.911}  & \textbf{0.748}  & \textbf{0.736} \\
\bottomrule[1pt]
\end{NiceTabular}
}
\vspace{-5pt}
\end{table}

\subsection{Comparison with Cross-Domain FIDL Methods}
To validate the effectiveness of our method in cross-domain forgery detection, we selected two mainstream models that claim to possess cross-domain detection capabilities, namely HiFi-Net~\cite{guo2023hierarchical} and FakeShield~\cite{xu2024fakeshield}, and evaluated them using their official weights. Additionally, we trained two mainstream image classification backbones, ResNet~\cite{he2016resnet} and CLIP~\cite{radford2021learning}, to serve as additional baselines.

We evaluated the detection performance of all models on the test sets of four major forgery detection tasks: CASIA1+, AIGCDetectionBench, DF40 (FR), and RTM-test. The results are shown in Table~\ref{table:cross}.
As can be seen from the table, UniShield significantly outperforms all baseline methods across all four tasks, demonstrating strong cross-domain generalization and detection robustness. Notably, in the DFD task, UniShield achieved an F1 score of 0.911, which is 0.201 higher than the second-best method FakeShield (F1 = 0.710), showcasing a substantial advantage. In the more challenging DMDL domain, UniShield also achieved an F1 score of 0.736, surpassing the best-performing ResNet (0.650) by approximately 0.086.

These results further highlight the limitations of existing methods: they fail to handle all categories of forgery types and show significant performance degradation when transferred to new domains. In contrast, UniShield provides end-to-end task-aware adaptation and a dynamic multi-expert model fusion mechanism, enabling robust detection of various forgery types within a unified framework, with superior practicality and generalization capability.

\begin{table}[t]
\centering
\caption{Comparison with IMDL expert methods on CASIA1+ and IMD2020 benchmarks. Img-F1 and Pix-F1 denote the F1 scores at the image level and pixel level.}
\label{table:IMDL}
\vspace{-8pt}
\renewcommand{\arraystretch}{1.2}
\resizebox{1.0\linewidth}{!}{  
\begin{NiceTabular}{c|cccc|cccc}
\CodeBefore
\tikz \fill [gray!10] (1-|1) rectangle (2-|10);   
\tikz \fill [gray!10] (2-|1) rectangle (3-|10); 
\Body
\toprule[1pt]
\multirow{2}{*}{Method} & \multicolumn{4}{c|}{CASIA1+} & \multicolumn{4}{c}{IMD2020} \\
& ACC & Img-F1 & IoU & Pix-F1 & ACC & Img-F1 & IoU & Pix-F1 \\
\midrule
SPAN        & 0.60 & 0.44 & 0.11 & 0.14 & 0.70 & 0.81 & 0.09 & 0.14 \\
MantraNet   & 0.52 & 0.68 & 0.09 & 0.13 & 0.75 & 0.85 & 0.10 & 0.16 \\
HiFi-Net    & 0.46 & 0.44 & 0.13 & 0.18 & 0.62 & 0.75 & 0.09 & 0.14 \\
PSCC-Net    & 0.90 & 0.89 & 0.36 & 0.46 & 0.67 & 0.78 & 0.22 & 0.32 \\
CAT-Net     & 0.88 & 0.87 & 0.44 & 0.51 & 0.68 & 0.79 & 0.14 & 0.19 \\
MVSS-Net    & 0.62 & 0.76 & 0.40 & 0.48 & 0.75 & 0.85 & 0.23 & 0.31 \\
FakeShield  & \underline{0.95} & \underline{0.95} & 0.54 & 0.60 & \underline{0.83} & \underline{0.90} & \underline{0.50} & \underline{0.57} \\
IML-ViT     & 0.92 & 0.93 & \underline{0.69} & \underline{0.76} & \underline{0.83} & \underline{0.90} & 0.30 & 0.39 \\
Ours        & \textbf{0.97} & \textbf{0.96} & \textbf{0.70} & \textbf{0.77} & \textbf{0.84} & \textbf{0.91} & \textbf{0.51} & \textbf{0.60} \\
\bottomrule[1pt]
\end{NiceTabular}
}
\vspace{-10pt}
\end{table}

\begin{table}[t]
\centering
\caption{Comparison with DeepFake detection expert methods on the DF40 benchmark. We report detection performance using the AUC metric on two datasets (FS and FR).}
\label{table:DeepFake}
\vspace{-8pt}
\renewcommand{\arraystretch}{1.2}
\setlength{\tabcolsep}{5pt}
\resizebox{\linewidth}{!}{%
\begin{NiceTabular}{c|cccccccc}
\CodeBefore
\tikz \fill [gray!10] (1-|1) rectangle (2-|10);   
\Body
\toprule[1pt]
Dataset & Xception & CLIP & SRM & SPSL & RECCE & RFM & DFD-R1 & Ours \\
\midrule
FS  & 0.991 & \underline{0.996} & 0.988 & 0.987 & 0.991 & 0.992 & 0.923 & \textbf{0.997} \\
FR  & 0.892 & \underline{0.908} & 0.867 & 0.849 & 0.855 & 0.884 & 0.897 & \textbf{0.913} \\
\bottomrule[1pt]
\end{NiceTabular}
}
\vspace{-5pt}
\end{table}

\begin{table}[t]
\centering
\caption{Comparison with DMDL expert methods on the RTM benchmark.}
\label{table:DMDL}
\vspace{-8pt}
\renewcommand{\arraystretch}{1.0}
\normalsize 
\resizebox{1.0\linewidth}{!}{%
\begin{NiceTabular}{c|ccccc}
\CodeBefore
\tikz \fill [gray!10] (1-|1) rectangle (2-|10);   
\Body
\toprule[1pt]
Method & IoU & Precision & Recall & Pix-F1 & Img-F1 \\
\midrule
UperNet       & 0.083  & 0.325 & 0.100  & 0.153 & 0.491 \\
DeepLabV3+    & 0.086  & 0.322 & 0.104 & 0.158 & 0.529 \\
HRNet-OCR     & 0.068  & 0.242 & 0.087  & 0.128 & 0.470 \\
SegFormer     & 0.157 & 0.384 & 0.210 & 0.272 & 0.615 \\
MaskFormer    & 0.137 & 0.260 & 0.225 & 0.241 & \underline{0.688} \\
Mask2Former   & 0.124 & 0.191 & 0.259 & 0.220 & 0.640 \\
RRU-Net       & 0.037  & 0.152 & 0.047  & 0.072  & 0.410 \\
PSCC-Net      & 0.033  & 0.036  & \textbf{0.303} & 0.064  & 0.687 \\
MVSS-Net++    & 0.051  & 0.073  & 0.144 & 0.097  & 0.549 \\
CAT-Net v2    & 0.113 & 0.302 & 0.153 & 0.203 & 0.548 \\
Liang et al.  & 0.046  & 0.054  & 0.250 & 0.088  & 0.596 \\
DTD           & 0.065  & 0.119 & 0.125 & 0.122 & 0.538 \\
ASC-Former    & \underline{0.197} & \underline{0.504} & 0.244 & \underline{0.329} & 0.633 \\
DMDL-R1        & 0.178 & 0.476 & 0.219 & 0.305 & 0.567 \\
Ours          & \textbf{0.209} & \textbf{0.521} & \underline{0.254} & \textbf{0.341} & \textbf{0.737} \\
\bottomrule[1pt]
\end{NiceTabular}%
}
\end{table}

\begin{table}[t]
\centering
\caption{Comparison with AIGC detection expert methods on the AIGCDetectBenchmark. We report detection accuracy across various generative models.}
\vspace{-8pt}
\label{table:AIGC}
\renewcommand{\arraystretch}{1.0}
\setlength{\tabcolsep}{1pt}
\scriptsize
\begin{NiceTabular}{c|cccccccc}
\CodeBefore
\tikz \fill [gray!10] (1-|1) rectangle (2-|10);   
\Body
\toprule[1pt]
Dataset & CNNSpot & FreDect & UnivFD & DIRE & PatchCraft & AIDE & FakeVLM & Ours \\
\midrule
ProGAN & \underline{1.000} & 0.994 & 0.998 & 0.528 & \textbf{1.000} & 1.000 & 0.997 & 0.999 \\
StyleGAN & 0.902 & 0.780 & 0.849 & 0.513 & 0.928 & \underline{0.996} & 0.917 & \textbf{0.998} \\
BigGAN & 0.712 & 0.820 & \underline{0.951} & 0.497 & \textbf{0.958} & 0.840 & 0.782 & 0.862 \\
CycleGAN & 0.876 & 0.788 & 0.983 & 0.496 & 0.702 & \underline{0.985} & 0.855 & \textbf{0.991} \\
StarGAN & 0.946 & 0.946 & 0.958 & 0.467 & \underline{1.000} & 0.999 & 0.991 & \textbf{1.000} \\
GauGAN & \underline{0.814} &0.806 & \textbf{0.995} & 0.512 & 0.716 & 0.733 & 0.749 & 0.783 \\
StyleGAN2 & 0.869 & 0.662 & 0.750 & 0.517 & 0.850 & \textbf{0.980} & \underline{0.958} & 0.973 \\
WFTIR & 0.917 & 0.508 & 0.869 & 0.533 & 0.822 & \underline{0.942} & 0.713 & \textbf{0.958} \\
ADM & 0.604 & 0.634 & 0.669 & \textbf{0.983} & 0.838 & 0.934 & 0.840 & \underline{0.948} \\
Glide & 0.581 & 0.541 & 0.625 & 0.924 & 0.901 & \underline{0.951} & 0.649 & \textbf{0.956} \\
Midjourney & 0.514 & 0.459 & 0.561 & \underline{0.895} & \textbf{0.954} & 0.772 & 0.660 & 0.835 \\
SD v1.4 & 0.506 & 0.388 & 0.637 & 0.912 & \underline{0.953} & 0.930 & 0.653 & \textbf{0.959} \\
SD v1.5 & 0.505 & 0.392 & 0.635 & 0.916 & 0.889 & \underline{0.929} & 0.850 & \textbf{0.939} \\
VQDM & 0.565 & 0.778 & 0.853 & 0.919 & 0.911 & \underline{0.952} & 0.673 & \textbf{0.961} \\
Wukong & 0.510 & 0.403 & 0.709 & 0.909 & \textbf{0.966} & \underline{0.936} & 0.810 & 0.946 \\
DALLE2 & 0.505 & 0.347 & 0.508 & 0.924 & 0.893 & \underline{0.966} & 0.864 & \textbf{0.970} \\
Mean & 0.708 & 0.640 & 0.784 & 0.715 & 0.893 & \underline{0.928} & 0.810 & \textbf{0.942} \\
\bottomrule[1pt]
\end{NiceTabular}
\vspace{-10pt}
\end{table}

\subsection{Comparison with Sub-Domain Expert Methods}

To demonstrate the effectiveness of UniShield across diverse forgery domains, we evaluate it on four major benchmarks, using the official weights of strong expert baselines. As shown in Tables~\ref{table:IMDL} to~\ref{table:AIGC}, UniShield consistently achieves top performance. 
In the IMDL task, as shown in Table~\ref{table:IMDL}, UniShield achieves strong performance on CASIA1+ and IMD2020. it reaches an image-level F1 score of 0.96 and a pixel-level F1 score of 0.77, outperforming the second-best model FakeShield, which achieves 0.95 and 0.60. In the DeepFake detection task, as Table~\ref{table:DeepFake} it obtains the highest AUC scores of 0.997 on the FS subset and 0.913 on the FR subset, exceeding all baselines including CLIP and DF-R1. For the DMDL domain, in Table~\ref{table:DMDL}, UniShield achieves the best image-level F1 score of 0.737 and pixel-level F1 score of 0.341, while also ranking first in precision and IoU. Finally, in AIGC detection, it achieves the highest average accuracy of 0.942 across 16 generative models. 
Notably, the cooperative reasoning mechanism in UniShield leads to better performance than any single model used within it. For example, in the IMDL task on CASIA1+, UniShield exceeds both FakeShield and IML-ViT, indicating the system’s ability to effectively integrate expert tools. This demonstrates a clear $1+1>2$ synergy, significantly boosting practical applicability, robustness, and cross-domain adaptability.


One example of our forgery detection report is shown in Figure~\ref{zhuguan}, based on an IMDL case. The report includes four components: a concise Description of the image content, Detection of tampering presence, Localization of the manipulated region, and a detailed Judgment Basis grounded in both low-level visual cues and high-level semantic inconsistencies. This structured and interpretable output demonstrates UniShield’s ability to not only detect and localize forgeries, but also to explain its reasoning in a way that enhances transparency and user trust.

\subsection{Robustness Study}

With the growing prevalence of the Internet and social media, individuals are more frequently exposed to images that suffer from transmission artifacts like JPEG compression and Gaussian noise. The performance of our model under these types of degradations is summarized in Table~\ref{table:robust}. This includes four common degradation scenarios: JPEG compression levels of 70 and 80, as well as Gaussian noise variances of 5 and 10.
Although UniShield experienced a slight performance decline under moderate disturbances, it remained within an acceptable range overall. This demonstrates the practicality and stability of our approach.






\begin{table}[h]
\centering
\caption{Robustness study. Our method maintains high robustness under various common degradations to different extents.}
\label{table:robust}
\vspace{-8pt}
\renewcommand{\arraystretch}{1.2}
\resizebox{1.0\linewidth}{!}{
\begin{NiceTabular}{l|cc|c|c|c}
\CodeBefore
\tikz \fill [gray!10] (1-|1) rectangle (2-|8);   
\tikz \fill [gray!10] (2-|1) rectangle (3-|8); 
\Body
\toprule[1pt]
\multirow{2}{*}{Degeneration} & \multicolumn{2}{c|}{IMDL} & AIGCD & DFD & DMDL \\
& ACC & F1 & ACC & AUC & F1 \\
\midrule
JPEG 70        & 0.868 & 0.871 & 0.871 & 0.85 & 0.639 \\
JPRG 80   & 0.914 & 0.919 & 0.895 & 0.864 & 0.668 \\
Gaussian 5      & 0.892 & 0.889 & 0.874 & 0.851 & 0.671 \\
Gaussian 10      & 0.847 & 0.851 & 0.852 & 0.819 & 0.642 \\
Original  & \textbf{0.971} & \textbf{0.963} & \textbf{0.942} & \textbf{0.913} & \textbf{0.736} \\
\bottomrule[1pt]
\end{NiceTabular}
}
\vspace{-15pt}
\end{table}

\subsection{Ablation Study}

 

To validate the effectiveness of our perception agent, we design four ablation variants as shown in Table~\ref{table:ablation}. The first two variants, case~(a): Always-LLM and case~(b): Always-non-LLM, disable the tool scheduler and instead always select either LLM-based or non-LLM-based tools, respectively. These settings help assess the importance of dynamic tool selection. The other two variants, case~(c): Any-vote and case~(d): Majority-vote, replace the perception agent with simple rule-based strategies for aggregating predictions from expert methods: Any-vote classifies an image as fake if any method detects it as fake, while Majority-vote does so only if at least half agree. These variants are evaluated under the same settings as UniShield, and the results are shown in Table~\ref{table:ablation}. Here, we report only the detection performance of each model. Our full method outperforms all variants across IMDL, AIGCD, DFD, and DMDL, demonstrating the advantages of both dynamic tool scheduling and learning-based integration of expert tools.

\begin{table}[h]
\centering
\caption{Ablation study. We compare different routing and tool selection strategies. Our full method achieves the highest performance across all domains.}
\label{table:ablation}
\vspace{-8pt}
\renewcommand{\arraystretch}{1.2}
\resizebox{1.0\linewidth}{!}{
\begin{NiceTabular}{c|l|cc|c|c|c}
\CodeBefore
\tikz \fill [gray!10] (1-|1) rectangle (2-|8);   
\tikz \fill [gray!10] (2-|1) rectangle (3-|8); 
\Body
\toprule[1pt]
\multirow{2}{*}{Case} & \multirow{2}{*}{Method} & \multicolumn{2}{c|}{IMDL} & AIGCD & DFD & DMDL \\
& & ACC & F1 & ACC & AUC & F1 \\
\midrule
(a)& Always-LLM        & 0.945 & 0.948 & 0.810 & 0.897  & 0.567 \\
(b)& Always-non-LLM   & 0.921 & 0.934 & 0.927 & 0.908 & 0.633 \\
(c)& Any-vote      & 0.597 & 0.569 & 0.510 & 0.660 & 0.594 \\
(d)& Majority-vote      & 0.781 & 0.748 & 0.832 & 0.751  & 0.603 \\
(e)& Ours           & \textbf{0.971} & \textbf{0.963} & \textbf{0.942} & \textbf{0.913} & \textbf{0.736} \\
\bottomrule[1pt]
\end{NiceTabular}
}
\vspace{-10pt}
\end{table}

\section{Conclusion}
In this study, we propose UniShield, a novel multi-agent forgery image detection system designed to handle all major forgery types in a unified manner. UniShield targets four primary manipulation domains, including IMDL, DMDL, DFD, and AIGCD, through a flexible and adaptive framework that supports task-aware routing and cross-domain generalization. The system consists of two collaborative agents: the perception agent, which identifies the forgery domain and dynamically selects appropriate tools, and the detection agent, which performs fine-grained forgery detection and generates structured, interpretable reports. UniShield integrates eight expert models spanning all four forgery domains, combining both LLM-based and non-LLM-based detection strategies. We conduct comprehensive evaluations on multiple authoritative datasets across four forgery sub-tasks, and results show that UniShield effectively integrates diverse expert models to achieve superior performance, consistently outperforming existing methods in both accuracy and robustness. 
In the context of rapidly advancing generative technologies, UniShield is expected to play a critical role in safeguarding the authenticity of visual content, with broad applicability in key areas such as forensic investigation, content verification, and portrait rights protection. Facing increasingly subtle and diverse forgeries, its scalable multi-agent design allows for the seamless and flexible integration of domain-specific expert tools to handle complex and evolving manipulations. Looking ahead, we expect UniShield as a vital component in building a trustworthy AI ecosystem, providing solid support for ensuring information security and maintaining public trust.

{
    \small
    \bibliographystyle{ieeenat_fullname}
    \bibliography{main}
}


\end{document}